\documentclass[final]{nesy2026}

\usepackage{microtype}
\usepackage{titlesec}
\titlespacing*{\section}{0pt}{8pt plus 2pt minus 2pt}{4pt plus 1pt minus 01pt}
\titlespacing*{\paragraph}{17pt}{0pt plus 0pt minus 0pt}{1em}
\title[Neuro-symbolic AI for Industrial Configuration]{Neuro-symbolic AI for Industrial Configuration}

 \clearauthor{
   \Name{Danilo Valerio} \Email{danilo.valerio@siemens.com}\\
   \Name{Philipp Kogler} \Email{philipp.kogler@siemens.com}\\
   \Name{Stefan Bischof} \Email{bischof.stefan@siemens.com}\\
   \addr Siemens AG Österreich%
   \AND
   \Name{Thomas Hubauer} \Email{thomas.hubauer@siemens.com}\\
   \Name{Huzefa Rangwala} \Email{huzefa.rangwala@siemens.com}\\
   \addr Siemens AG
 }

\begin{document}

\maketitle

\begin{abstract}
Large Language Models (LLMs) have shown impressive performance on a wide range of generative tasks. Yet their probabilistic nature makes them, in isolation, fundamentally unsuited for industrial \emph{product configuration}, where outputs must be syntactically valid, semantically consistent with a knowledge base of hundreds of features and rules, and producible by an existing manufacturing chain. We argue that \emph{Neuro-symbolic (NeSy) AI} methods lay out a promising path towards industrial-grade configurators that are reliable by design, explainable, and trustworthy. This paper describes a taxonomy of three NeSy integration strategies, namely hybrid inference, hybrid fine-tuning, and hybrid training, exploring their usage in the configuration domain. %
We report our effort to operationalize NeSy concepts in an industrial configuration copilot and derive a set of practical design choices for deploying trustworthy AI in engineering environments. We close with a discussion of open research challenges we consider most pressing, in particular how to scale NeSy methods from small academic demonstrators to the size of industrial configurators.
\end{abstract}

\section{Introduction}
\label{sec:intro}
Industrial configurators are AI-enabled decision-support systems that transform customer requirements into valid, consistent, and producible product or system specifications while guaranteeing compliance with domain knowledge, engineering constraints, and operational requirements. Industrial configurators have been used successfully in practice for decades across domains such as power generation, factory automation, energy systems, and mobility. In these domains, correctness, traceability, and compliance with strict domain rules are not optional features but contractual and regulatory obligations. For this reason, traditional knowledge-based configurators are based on symbolic AI techniques such as rule-based reasoning, knowledge representation, and constraint satisfaction~\citep{falkner2016twentyfive}. In a typical workflow, an engineer incrementally specifies product features and requirements, while a symbolic engine ensures that every decision remains consistent with the domain model and does not violate any constraints~\citep{falkner2020solver}.

The recent rise of LLMs has created both an opportunity and a temptation to make these configurators ``talk''. Natural-language-based configuration copilots could substantially lower the entry barrier for engineers, shorten Configure-Price-Quote (CPQ) cycles, and transform a configurator %
into a conversational interface ~\citep{lawless2024, kogler2024configcopilot, Kogler2026configGenAI}. Our internal evaluations have however shown that configurators relying solely on LLMs fail to satisfy the stringent reliability requirements of industrial engineering for the following four reasons: \textit{(i)} Syntactic hallucinations, where the model invents properties, attributes, or concepts that do not exist in the product model; \textit{(ii)} Semantic hallucinations, where the model assigns invalid values to existing features or violates their semantic constraints; \textit{(iii)} Producibility violations, where the proposed configuration is internally inconsistent, violates domain rules, and cannot be manufactured; %
\textit{(iv)} Intent misalignment, where the generated configuration is valid, yet fails to satisfy the user's actual requirements and preferences.

Beyond these practical shortcomings lies a broader challenge: industrial AI ultimately depends on vast amounts of structured domain knowledge accumulated over decades. As the benefits of simply scaling data and compute begin to plateau, leveraging this knowledge explicitly will become a key differentiator for next-generation AI systems. We therefore argue that neuro-symbolic AI represents the most promising path towards industrial-grade configurators, combining the pattern-matching, language-understanding, and generalization capabilities of neural networks with the explicit knowledge representation, constraint satisfaction, and formal correctness guarantees provided by symbolic AI~\citep{garcez2020thirdwave,shakarian2023nesybook}. This view is increasingly reflected in the broader AI community, where hybrid approaches are widely regarded as a key direction for overcoming the limitations of purely neural and purely symbolic systems.

\section{Neuro-Symbolic Industrial Configurators}
\label{sec:methods}

The industrial configuration task we address takes as input a natural-language requirement (a free-text description of the desired product, possibly under-specified or ambiguous) and an encoding of the product model, containing \textit{(i)} the finite set of product configuration variables, \textit{(ii)} the legal values for these variables, and \textit{(iii)} a set of hard constraints across variables, whose violation makes a configuration not producible, e.g., compatibility rules, include/exclude rules, default rules. The system is judged \emph{industrial-grade} on this task if it produces as output a configuration that simultaneously satisfies four properties:
\vspace{-5pt}
\begin{enumerate}
  \item \textbf{Syntactic correctness}~$\Phi_{\text{syn}}$: Syntax is respected and all variables are valid.\vspace{-8pt}
  \item \textbf{Semantic correctness}~$\Phi_{\text{sem}}$:  All values assigned to the variables are legal.\vspace{-8pt}
  \item \textbf{Producibility}~$\Phi_{\text{prod}}$: All hard producibility constraints are satisfied.\vspace{-8pt}
  \item \textbf{Intent matching}~$\Phi_{\text{int}}$: The resulting product satisfies the user's requirements.%
\end{enumerate}
\vspace{-5pt}
Approaches relying solely on LLMs address $\Phi_{\text{int}}$ well but increasingly fail on $\Phi_{\text{syn}}$, $\Phi_{\text{sem}}$, and $\Phi_{\text{prod}}$
as the size of the product model and the number/complexity of constraints grows~\citep{shojaee2025illusion}. Pure symbolic approaches typically guarantee $\Phi_{\text{syn}}$, $\Phi_{\text{sem}}$, and $\Phi_{\text{prod}}$
but require the user to express requirements in a formal modeling language or GUI. The position of this paper is therefore that the research goal for industrial configuration is to design neuro-symbolic architectures that achieve \emph{all four} properties \emph{simultaneously}.

\begin{figure}
    \centering
    \vspace{-20pt}
    \includegraphics[width=1\linewidth]{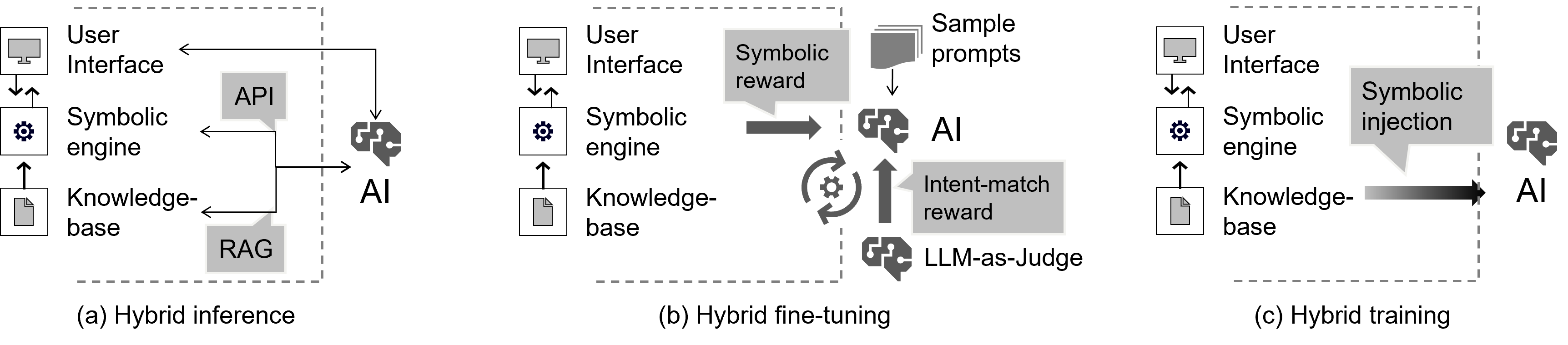}
    \vspace{-25pt}
    \caption{Three neuro-symbolic integration strategies for industrial configuration. (a) Hybrid inference constrains generation at decode time. (b) Hybrid fine-tuning uses symbolic feedback as a training reward signal. (c) Hybrid training embeds constraints directly into the model architecture}
    \vspace{-24pt}
    \label{fig:methods}
\end{figure}

We view neuro-symbolic AI as a layered integration of neural and symbolic techniques~\citep{kautz2022} and distinguish three complementary approaches (depicted in~\figureref{fig:methods}), characterized by where symbolic knowledge is introduced into the neural pipeline.

\paragraph{Hybrid inference}
\label{sec:hybrid-inference}
\hspace{-10pt} couples a pre-trained LLM with a symbolic reasoning engine that actively constrains generation. The symbolic component operates over a formal knowledge base organized into three layers: \textit{(i)} a structural schema defining the exchange format (e.g., JSON), \textit{(ii)} a vocabulary of legal configuration elements such as features, attributes, and components, and \textit{(iii)} a set of constraints capturing their admissible relationships.

In our reference architecture, reasoning is performed at two levels of granularity. At the lowest level, a \emph{token reasoner} performs grammar-constrained decoding~\citep{geng2024grammarconstraineddecoding} by filtering the next-token distribution of the LLM at every decoding step and ensuring that only tokens compatible with the schema and the partially generated output remain admissible. This mechanism prevents the generation of invalid properties, values, or structures. At a higher level, an \emph{element reasoner} is invoked whenever a configuration element is completed. Using constraint propagation and look-ahead reasoning, it evaluates the consequences of the partial configuration restricting subsequent generation to the valid subset of elements~\citep{kogler2026nesydecoding}. For example, selecting an electric motor may require the specification of input voltage and torque before generation can proceed.

Hybrid Inference currently represents the most mature neuro-symbolic paradigm for industrial copilots. It requires no retraining of the foundation model, can be integrated with existing configuration engines, and provides strong guarantees by construction: since correctness is delegated entirely to the symbolic layer, the solver guarantees that generated configurations do not contain hallucinated features or values and remain consistent with all domain constraints. Its primary limitation is the additional inference-time complexity introduced by repeated interactions between the neural and symbolic components. %

\paragraph{Hybrid fine-tuning}
\label{sec:hybrid-finetuning}
\hspace{-10pt} shifts the interaction between neural and symbolic components from inference time to training time. Instead of constraining generation through a reasoning engine, the symbolic component is used to provide feedback that guides the adaptation of the model parameters. The objective is to make the LLM internalize domain regularities to produce more reliable output in the absence of explicit symbolic intervention at inference. The central idea is to place a symbolic solver in the training loop. Candidate configurations generated by the neural model are evaluated against the knowledge base and assigned a reward according to their degree of correctness. Solutions that satisfy domain constraints receive positive reinforcement, whereas configurations containing invalid features, inconsistent values, or rule violations are penalized. The resulting feedback signal can be incorporated with techniques such as Reinforcement Learning with Symbolic Feedback~\citep{jha2025rlsf} or Reinforcement Learning with Verifiable Rewards~\citep{lambert2025tulu}.

A reward function based solely on symbolic feedback maximizes the likelihood of producing valid configurations. However, optimizing exclusively for validity can lead to reward-hacking behaviors. For example, the model may converge towards repeatedly generating a small set of highly rewarded configurations regardless of the user's requirements. To mitigate this issue, we decompose the reward function as $R_{\mathrm{total}}=w_1\left(R_{\mathrm{feature}}\right)+w_2\left(R_{\mathrm{config}}\right)+w_3\left(R_{\mathrm{intent}}\right)$
where $R_{\mathrm{feature}}$ measures the validity of individual feature assignments (univariate feedback), $R_{\mathrm{config}}$ measures the validity of the complete configuration (multivariate feedback), and $R_{\mathrm{intent}}$ measures the degree to which the generated recommendation satisfies user's requirements. The first two terms are derived from symbolic verification by the constraint solver, while the latter is estimated using an LLM-as-a-judge that evaluates requirement satisfaction from the natural-language interaction.

Compared to hybrid inference, this approach transfers part of the symbolic knowledge into the model parameters themselves. As a consequence, inference becomes significantly faster because the model no longer needs to repeatedly query an external solver. At the same time, the approach can improve the semantic validity of generated configurations and reduce the frequency of hallucinations. However, correctness guarantees become statistical rather than deterministic: the model learns to prefer valid configurations, but cannot guarantee constraint satisfaction for every output. In safety-critical domains, symbolic validation may therefore still be required as a final verification step.

\paragraph{Hybrid training}
\label{sec:hybrid-training}
\hspace{-10pt} embeds symbolic knowledge directly into the neural network architecture or training objective, such that constraints are enforced \emph{by construction}. While this approach is currently impractical for LLM-scale models due to scalability and computational challenges, it remains highly relevant for industrial configurators.
Our exemplar is a \emph{compiled neuro-symbolic recommender system}. Historical configuration data are represented as a user--configuration matrix, where rows correspond to customers and columns to product features selected in previous configurations. Recommendation then becomes a constrained matrix-completion problem. We address this problem using a Neural Collaborative Filtering (NCF) model~\citep{ww2017}, augmented with a Logic Tensor Network (LTN) regularizer~\citep{badreddine2022ltn} that compiles domain constraints into differentiable loss terms. During training, the model is therefore optimized not only to reconstruct historical preferences, but also to satisfy engineering rules encoded in the knowledge base.

\section{Discussion}
\label{sec:results}
The three methodologies presented in this paper should be viewed as complementary rather than competing. We envision future industrial configurators combining all three layers: hybrid training for recommendation and decision-support components, hybrid fine-tuning to internalize frequently occurring domain rules and reduce inference costs, and hybrid inference as an outer safety layer providing hard correctness guarantees during user interaction. While the conceptual foundations are increasingly well understood, their application to industrial-scale configuration remains in its infancy, and we are currently conducting preliminary experiments on industrial product models to better understand the practical trade-offs among reliability, latency, explainability, and implementation effort. 

Important research challenges remain. First, \emph{scalability}: most neuro-symbolic approaches have been validated on relatively small academic benchmarks, whereas industrial configurators routinely involve millions of variables, constraints, and engineering artifacts. Developing architectures that preserve symbolic guarantees at such scales remains an open problem. Closely related is the question of \emph{knowledge allocation}: determining which forms of domain knowledge should remain explicitly represented and which can be safely internalized through fine-tuning or training may ultimately determine the practicality of large-scale industrial neuro-symbolic systems. Finally, \emph{evaluation methodology}: current studies employ heterogeneous datasets and metrics, making rigorous comparison difficult. Building on initiatives such as \textsc{IndusCP}~\citep{shi2025constraintllm} and \textsc{JsonSchemaBench}~\citep{geng2025jsonschemabench}, the field would benefit from a shared benchmark for neuro-symbolic configuration. Such a benchmark could play a role analogous to \textsc{HumanEval} in code generation, providing common tasks, evaluation protocols, and reproducible baselines that accelerate progress. %

\bibliography{nesy2026-valerio}
\end{document}